\documentclass{article}
\usepackage{spconf,amsmath,epsfig}
\usepackage{cite}
\usepackage{multirow}
\usepackage{amsmath}
\usepackage{amsfonts}
\usepackage[colorlinks,
            linkcolor=red,
            anchorcolor=blue,
            citecolor=blue,
            ]{hyperref}
            
\let\OLDthebibliography\thebibliography
\renewcommand\thebibliography[1]{
  \OLDthebibliography{#1}
  \setlength{\parskip}{0pt}
  \setlength{\itemsep}{0pt plus 0.3ex}
}
\usepackage{xspace}
\usepackage{color}
\usepackage{multirow}
\usepackage{graphics}
\usepackage{adjustbox}
\usepackage{subfigure}
\usepackage{paralist} % compactitem, compactenum
\usepackage{enumitem} % adjust enum margin
\usepackage{booktabs} % Publication quality tables
\usepackage{array}
\usepackage{caption} % captionof
\usepackage[ruled,linesnumbered,boxed]{algorithm2e}
\graphicspath{{figures/}}

\definecolor{gray}{rgb}{0.35,0.35,0.35}
\definecolor{MyBlue}{rgb}{0,0.2,0.8}
\definecolor{MyRed}{rgb}{0.8,0.2,0}
\definecolor{MyGreen}{rgb}{0.0,0.4,0.1}
\definecolor{MyGray}{rgb}{0.4,0.4,0.4}

\long\def\ignorethis#1{}

\newlength\paramargin
\newlength\figmargin
\newlength\secmargin

\newcolumntype{L}[1]{>{\raggedright\let\newline\\\arraybackslash\hspace{0pt}}m{#1}}
\newcolumntype{C}[1]{>{\centering\let\newline\\\arraybackslash\hspace{0pt}}m{#1}}
\newcolumntype{R}[1]{>{\raggedleft\let\newline\\\arraybackslash\hspace{0pt}}m{#1}}

\def\eg{\textit{e.g.},\xspace}

\def\etal{\textit{et~al.}\xspace}

\newcommand{\secref}[1]{Section~\ref{sec:#1}}
\newcommand{\figref}[1]{Fig.~\ref{fig:#1}}
\newcommand{\tabref}[1]{Table~\ref{tab:#1}}

\begin{document}\sloppy

% Example definitions.
% --------------------
\def\x{{\mathbf x}}
\def\L{{\cal L}}

% Title.
% ------
\title{Unifying Generation and Compression: 
\\
Ultra-low bitrate Image Coding Via Multi-stage Transformer
\thanks{*The corresponding author.}
}
%
% Single address.
% ---------------
% \name{Naifu Xue$^{\ast}$, Qi Mao$^{\ast}$, Zijian Wang$^{\ast}$, Yuan Zhang$^{\ast}$, Siwei Ma$^{\dag}$}
% \address{$^{\ast}$Communication University of China, $^{\dag}$Peking University}
\name{Naifu Xue$^{1}$, Qi Mao$^{1*}$, Zijian Wang$^{1}$, Yuan Zhang$^{1}$, Siwei Ma$^{2}$}
\address{$^{1}$Communication University of China, $^{2}$Peking University}

\maketitle

\begin{abstract}
Recent progress in generative compression technology has significantly improved the perceptual quality of compressed data. 
However, these advancements primarily focus on producing high-frequency details, often overlooking the ability of generative models to capture the prior distribution of image content, thus impeding further bitrate reduction in extreme compression scenarios ($<0.05$ bpp).
Motivated by the capabilities of predictive language models for lossless compression, this paper introduces a novel \emph{Unified Image Generation-Compression} (UIGC) paradigm, merging the processes of generation and compression.
A key feature of the UIGC framework is the adoption of vector-quantized (VQ) image models for tokenization, alongside a multi-stage transformer designed to exploit spatial contextual information for modeling the prior distribution.
As such, the dual-purpose framework effectively utilizes the learned prior for entropy estimation and assists in the regeneration of lost tokens. 
Extensive experiments demonstrate the superiority of the proposed UIGC framework over existing codecs in perceptual quality and human perception, particularly in ultra-low bitrate scenarios ($\leq 0.03$ bpp), pioneering a new direction in generative compression.

\end{abstract}
\begin{keywords}
Generative Compression, Extreme Compression, Image Generation, VQGANs, Transformer
\end{keywords}

%%%%%%%%%%%%%%%%%%%%%%%%%%%%%%%%%%%%%%%%%%%%%%%%%%%%%%%%%%%%%%%%%%%%%%%%%%
% Intro
%%%%%%%%%%%%%%%%%%%%%%%%%%%%%%%%%%%%%%%%%%%%%%%%%%%%%%%%%%%%%%%%%%%%%%%%%%
\section{Introduction}
\label{sec:intro}

%%%%%%%%%%%%%%%%%%%%%%%%%%%%%%%%%%%%%%%%%%%%%%%%%%%%%%%%%%%%%%%%%%%%%%%%%%
% para1: 
% Background, ultra-low bitrate compression
% drawbacks of existing compression method
%%%%%%%%%%%%%%%%%%%%%%%%%%%%%%%%%%%%%%%%%%%%%%%%%%%%%%%%%%%%%%%%%%%%%%%%%%

%%%%%%  current version  %%%%%%
Ultra-low bitrate compression presents a significant challenge in the field of image/video compression, particularly due to substantial information loss when faced with extremely limited network bandwidth, such as in satellite communications.
Traditional block-based compression codecs, \eg VVC~\cite{vvc}, are constrained to use large quantization steps in such scenarios, inevitably leading to noticeable blurring and blocking artifacts.
Despite the superior rate-distortion (R-D) performance of learning-based compression techniques~\cite{NEURIPS2018_53edebc5, Cheng_2020_CVPR, He_2021_CVPR_checkerboard, lu2022highefficiency}, these methods produce blurry images at low bitrates, due to the reliance on optimization of pixel-oriented distortion metrics.

%%%%%%  current version  %%%%%%
%
To address challenges in ultra-low bitrate scenarios, generative compression methods~\cite{Agustsson_2019_ICCV, NEURIPS2020_8a50bae2, fcc, Agustsson_2023_CVPR, mao_layerd, mao_conceptual, 2021thousand_to_one, chang2023semantic, mao2023extreme} have employed generative models \cite{kingma2022autoencoding, goodfellow2014generative, ho2020denoising} to enhance the visual quality of decoded images, with a focus primarily on the generator's ability to \emph{produce high-frequency details}.
This paradigm follows two primary technical pathways:
one involves training existing end-to-end image codecs using perceptual and adversarial losses~\cite{Agustsson_2019_ICCV, NEURIPS2020_8a50bae2, fcc, Agustsson_2023_CVPR}, and the other~ \cite{mao_layerd, mao_conceptual, 2021thousand_to_one, chang2023semantic, mao2023extreme} leverages specially designed encoders to compress images into more compact representations.
However, despite their effectiveness, these methods tend to overlook \emph{modeling the prior distribution of image content}, a critical aspect that differentiates image generation from image reconstruction task.
In situations where significant information loss occurs due to extremely limited bandwidth, it is plausible to reconstruct some of the lost content by sampling from the prior distribution.

\begin{figure}[!t]
    \centering
    \includegraphics[width=1.0\linewidth]{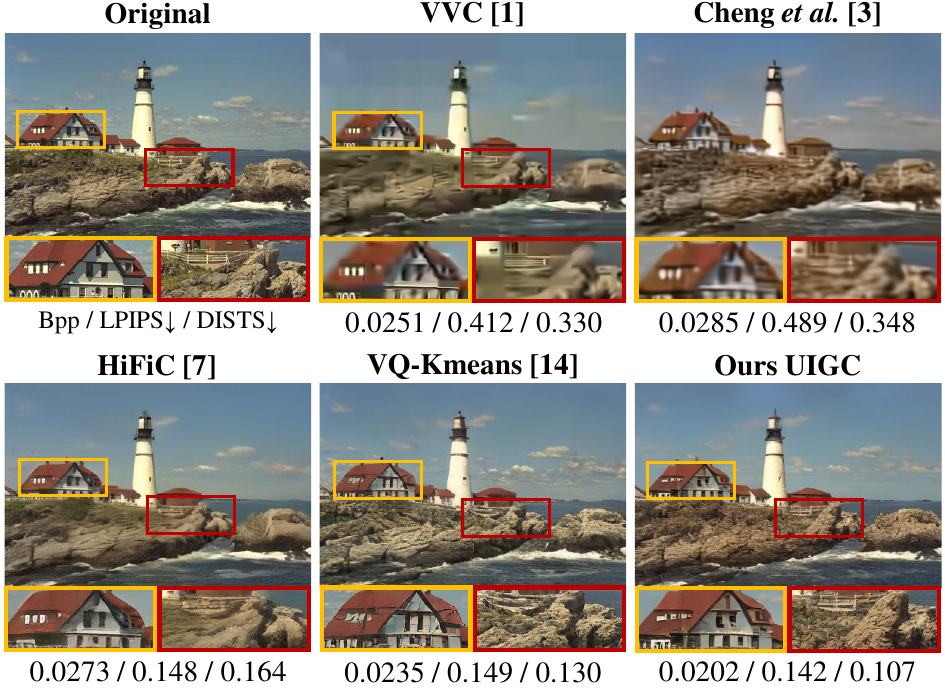}
    \caption{\textbf{Qualitative comparisons between state-of-the-art image compression methods}, including traditional \cite{vvc}, learning-based \cite{Cheng_2020_CVPR}, generative-based\cite{NEURIPS2020_8a50bae2, mao2023extreme}, and Ours.}
    \label{fig:first}
\end{figure}

%%%%%%%%%%%%%%%%%%%%%%%%%%%%%%%%%%%%%%%%%%%%%%%%%%%%%%%%%%%%%%%%%%%%%%%%%%
% para3: 
% generation-compression qeuivalence
% unified image gereration-compression, UIGC
%%%%%%%%%%%%%%%%%%%%%%%%%%%%%%%%%%%%%%%%%%%%%%%%%%%%%%%%%%%%%%%%%%%%%%%%%%

\begin{figure*}
    \centering    \includegraphics[width=1.0\linewidth]{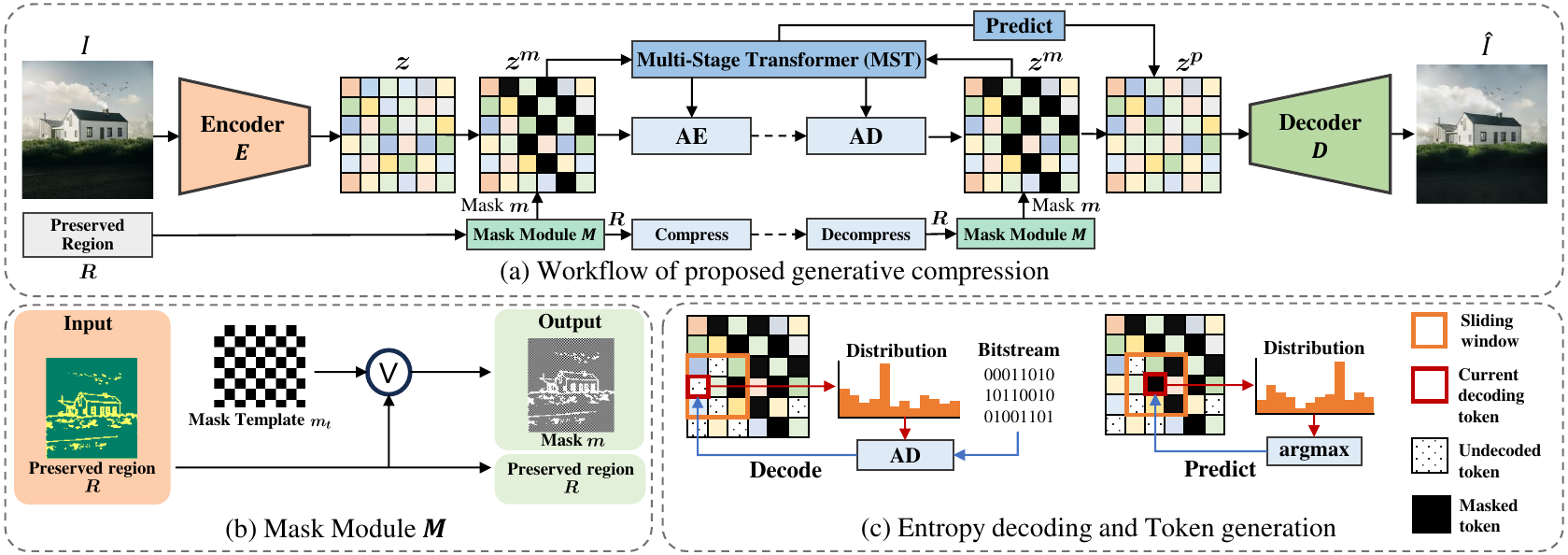}
    \caption{\textbf{Overview of the proposed UIGC framework.} (a) the overall compression workflow: we adopt a multi-stage transformer, and AE/AD denote arithmetic encoder/decoder; (b) the mask mechanism using mask module $\boldsymbol{M}$, and $\lor$ denotes the logical ``OR" operator; (c) entropy decoding and token generation on the decoder side.}
    \label{fig:overview}
\end{figure*}

Meanwhile, the fundamental aspect of entropy estimation requires accurately determining the prior probability distribution of symbols, thereby boosting the efficiency of entropy coding.
Consequently, the mathematical equivalence in estimating this \emph{prior probability distribution}—between entropy minimization in lossless compression and log$_2$-likelihood maximization in generation—poses a critical inquiry:
\emph{Is it possible to develop a method that models the prior distribution for both entropy estimation in compression and sampling in generation, all within a cohesive and unified framework?}
Recently, in the field of natural language process (NLP), Del{\'e}tang \etal\cite{deletang2023language}  have demonstrated that sequence generation models, such as large language models (LLMs), can be effectively used for lossless compression. 
Nevertheless, the extensive representation space of images presents a significant challenge in efficiently modeling the prior distribution.
On the brighter side, advancements in Vector-Quantized Image Modeling (VIM)~\cite{Esser_2021_CVPR, Chang_2022_CVPR} have made strides in compressing images into compact and discrete token representations using the vector-quantized (VQ) encoder. 
This development paves the way for transforming images into compact and discrete token representations via VIM, which enables the utilization of discrete generative models, similar to LLMs, for both entropy estimation and token generation.
In this work, we present a novel image compression paradigm, the Unified Image Generation-Compression (UIGC) framework, innovatively designed to facilitate \emph{both entropy encoding of tokens} and \emph{the prediction of lost tokens}. 
By converting images into discrete token representations using VIM \cite{Esser_2021_CVPR}, 
our UIGC codec focuses on accurate prior modeling of these tokens and the strategic discarding of nonessential tokens, leading to enhanced bitrate reduction while still producing perceptually pleasing images.
Departing from the traditional autoregressive \cite{Esser_2021_CVPR} and non-autoregressive \cite{Chang_2022_CVPR} models typically used in NLP, we propose a \textbf{Multi-Stage Transformer} (MST) specifically tailored to image characteristics. 
The MST restructures the autoregressive order by dividing the token map into four groups, enabling most tokens to effectively utilize the surrounding context for prediction.
Recognizing the crucial role of structural information in visual perception, we incorporate an edge-preserved checkerboard mask pattern, which selectively discards tokens while maintaining essential structural details.
With MST's multi-stage order, the prediction of lost tokens is significantly enhanced, utilizing the surrounding content to ensure the generation of high-quality images.
To evaluate the efficiency of the proposed UIGC framework, we conduct experiments on the Kodak \cite{kodak} and CLIC \cite{CLIC2020} datasets.
The experimental results, both quantitative and qualitative, demonstrate that our method surpasses existing techniques in maintaining perceptual quality under ultra-low bitrate conditions \textbf{($\leq$ 0.03 bpp)}.
%
% As shown in \figref{first}, our framework effectively reduces the bitrate while maintaining compromising perceptual image quality.
As shown in \figref{first}, our framework effectively reduces the bitrate while maintaining uncompromising image quality.
%

%%%%%%%%%%%%%%%%%%%%%%%%%%%%%%%%%%%%%%%%%%%%%%%%%%%%%%%%%%%%%%%%%%%%%%%%%%
% Method
%%%%%%%%%%%%%%%%%%%%%%%%%%%%%%%%%%%%%%%%%%%%%%%%%%%%%%%%%%%%%%%%%%%%%%%%%%
\section{Unifying Generation and Compression}
\label{sec:method}

%%%%%%%%%%%%%%%%%%%%%%%%%%%%%%%%%%%%%%%%%%%%%%%%%%%%%%%%%%%%%%%%%%%%%%%%%%
% Overview
% Describe the workflow
% Describe the entropy coding and token generation process
%%%%%%%%%%%%%%%%%%%%%%%%%%%%%%%%%%%%%%%%%%%%%%%%%%%%%%%%%%%%%%%%%%%%%%%%%%
In this work, we aim to compress the image $I$ at ultra-low bitrates while reconstructing the image $\hat{I}$ with a pleasing perceptual quality. 
In contrast to previous generative compression approaches that predominantly concentrate on the reconstruction of high-frequency details, our proposed UIGC framework shifts its emphasis toward modeling the prior distribution of image content for both \emph{entropy estimation} and \emph{content generation}.
\figref{overview} presents the overview of the proposed method.
In the following sections, we provide a detailed explanation of the image coding methodologies in \secref{coding_workflow}. 
Subsequently, we introduce the proposed Multi-Stage Transformer (MST) in \secref{mst} and the edge-preserved checkerboard mask in \secref{mask} to leverage spatial contextual dependencies in image token maps and effectively eliminate redundant tokens.
This approach facilitates efficient prior modeling and bitrate savings while preserving image quality.

\begin{figure}[ht]
    \centering
    \includegraphics[width=1.0\linewidth]{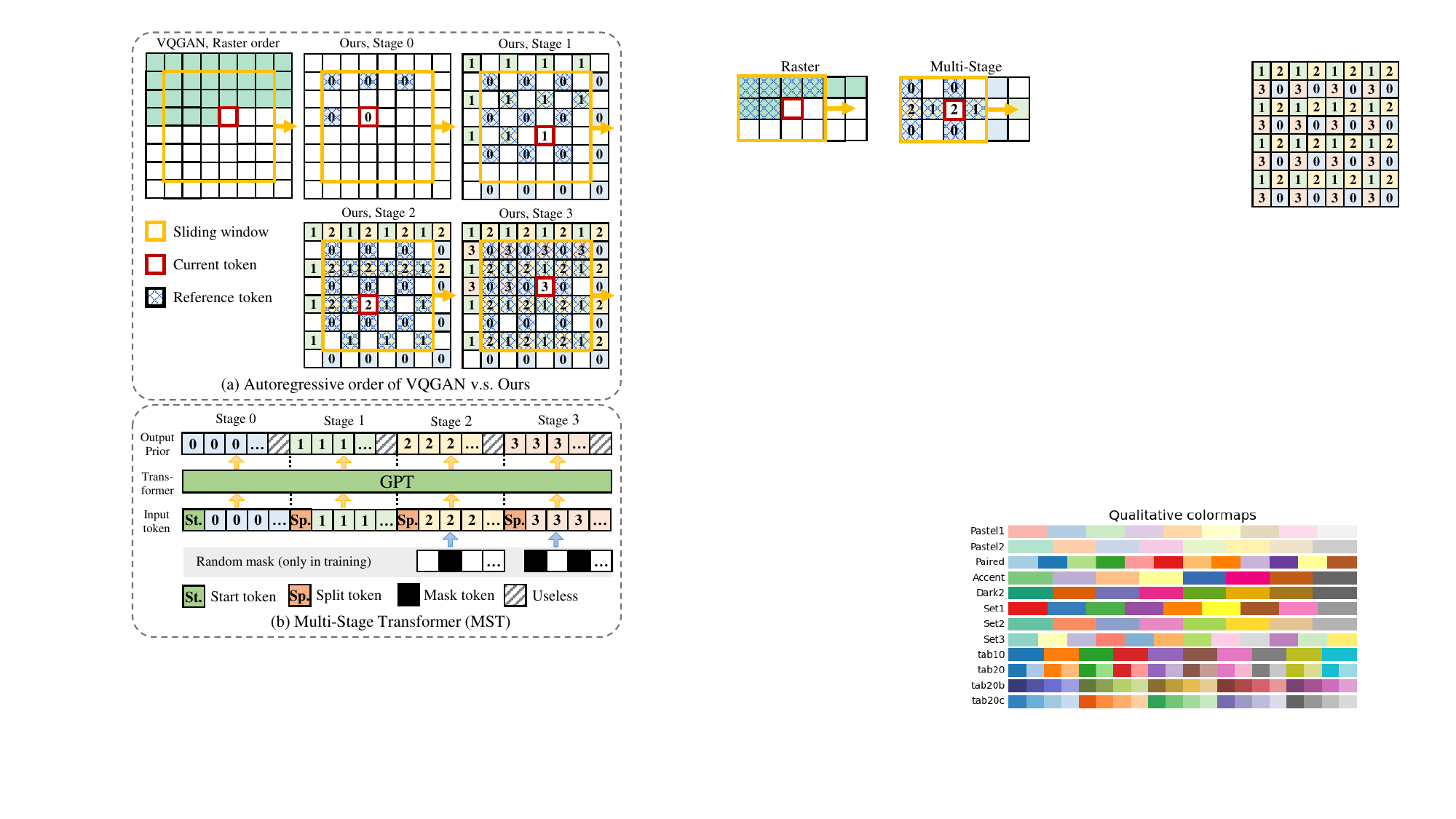}
    \caption{\textbf{Our MST.} 
    %
    % The 6x6 sliding window is for illustration only. 
    %
    We implement a GPT-style transformer for the prior modeling at each stage.}
    \label{fig:transformer}
\end{figure}

\subsection{Image Coding via UIGC}
\label{sec:coding_workflow}
The primary goal of our UIGC framework is to leverage the prior distribution for both entropy estimation and content generation. 
To achieve this, we selectively compress essential tokens, discarding redundant ones to achieve bitrate savings. 
The discarded tokens are then generated directly on the decoder side.
As illustrated in \figref{overview}(a), on the encoder side, the VQ encoder $\boldsymbol{E}$ \cite{Esser_2021_CVPR} transforms the given image $I$ into a token representation $\boldsymbol{z}\in \mathbb{N}^{h\times w}$. 
Then, we discard redundant tokens using the mask mechanism in \secref{mask}, where the mask $\boldsymbol{m}\in \{0,1\}^{h \times w}$ replaces discarded tokens as $[\text{Mask}]$:

\begin{equation}
    z^{m}_{i,j} = m_{i,j} \, z_{i,j} + (1-m_{i,j})  [\text{Mask}]. \quad i \in h, j \in w
    \label{eq:add_mask}
\end{equation}

Subsequently, the masked token map $\boldsymbol{z}^m$ is compressed by the arithmetic compression codec using the prior distribution of the MST, while the $[\text{Mask}]$ tokens are skipped.
Since the mask $\boldsymbol{m}$ is generated by our Mask Module $\boldsymbol{M}$ according to the preserved region $\boldsymbol{R}$, we further losslessly compress this region and transmit it to the decoder side.
On the decoder side, we restore the layout of the token map $\boldsymbol{z}^{m}$ using the decoded mask $\boldsymbol{m}$.
As illustrated in \figref{overview}(c), the unmasked tokens undergo decompression using the prior distribution from the MST.
Simultaneously, the masked tokens are predicted with the highest probability from this prior distribution.
Thus, the $\boldsymbol{z}^{p}$ is derived as:
\begin{equation}
    z^{p}_{i,j} = m_{i,j} \, z^{m}_{i,j} + (1-m_{i,j})  \,  ({\operatorname{argmax}}\, p_{i,j}).
    \label{eq:pred_mask}
\end{equation}
Finally, the VQ decoder $\boldsymbol{D}$ \cite{Esser_2021_CVPR} utilizes $\boldsymbol{z}^{p}$ to reconstruct the decoded image $\hat{I}$.

%%%%%%%%%%%%%%%%%%%%%%%%%%%%%%%%%%%%%%%%%%%%%%%%%%%%%%%%%%%%%%%%%%%%%%%%%%
% transformer module
% raster order transformer, unable to utilize the token in lower right, fig3a
% checkerboard mask, masked token are surrounded by available tokens, if can utilize those information
% based on this, we design a MST, which changes the autoregressive order by spatial grouping
% as fig 3b, 4 groups, 
%%%%%%%%%%%%%%%%%%%%%%%%%%%%%%%%%%%%%%%%%%%%%%%%%%%%%%%%%%%%%%%%%%%%%%%%%%

\begin{figure*}[!t]
    \centering
    \includegraphics[width=1.0\linewidth]{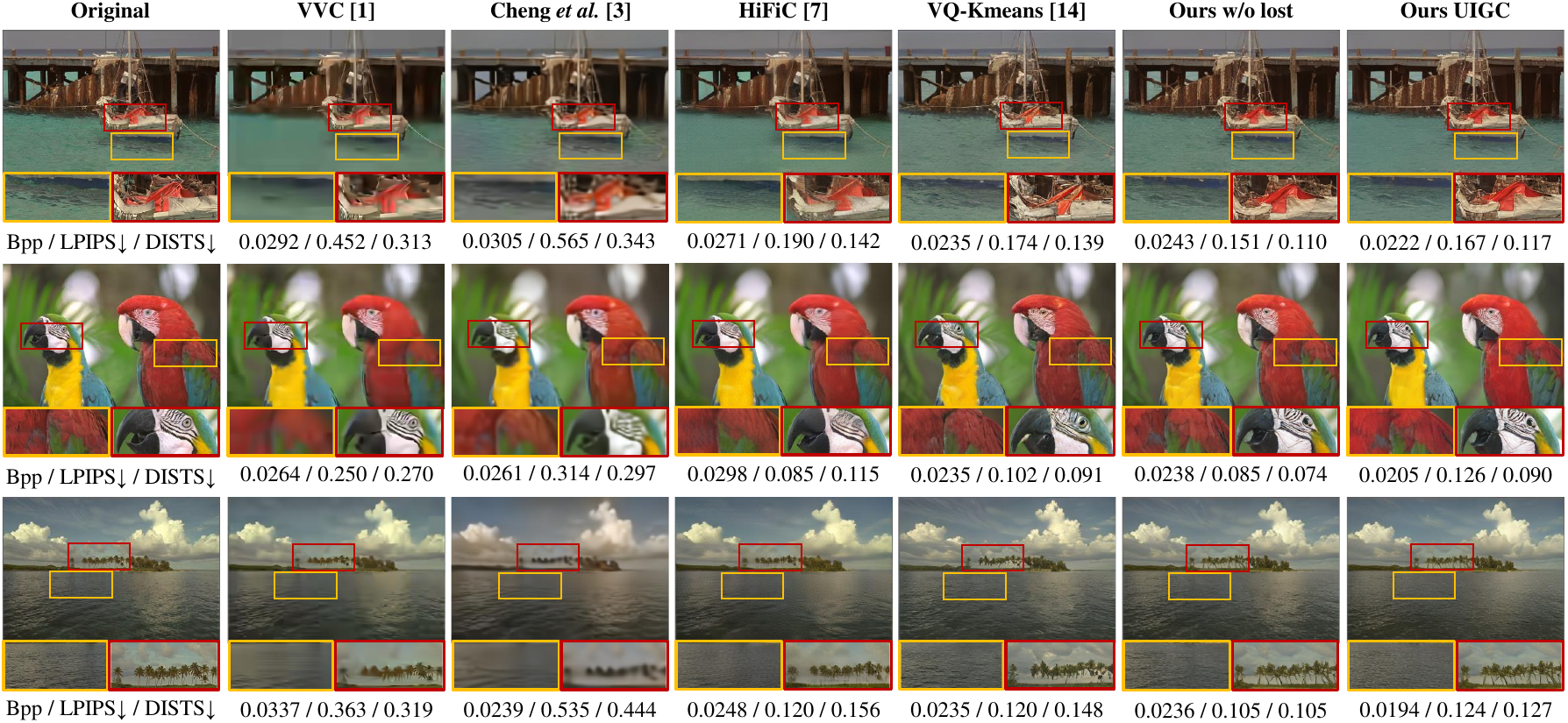}
    \caption{\textbf{Qualitative comparisons on the Kodak dataset \cite{kodak}.} In particular, ↓ indicates that lower is better.}
    \label{fig:visual_kodak}
\end{figure*}

\begin{figure*}[!t]
    \centering
    \includegraphics[width=1.0\linewidth]{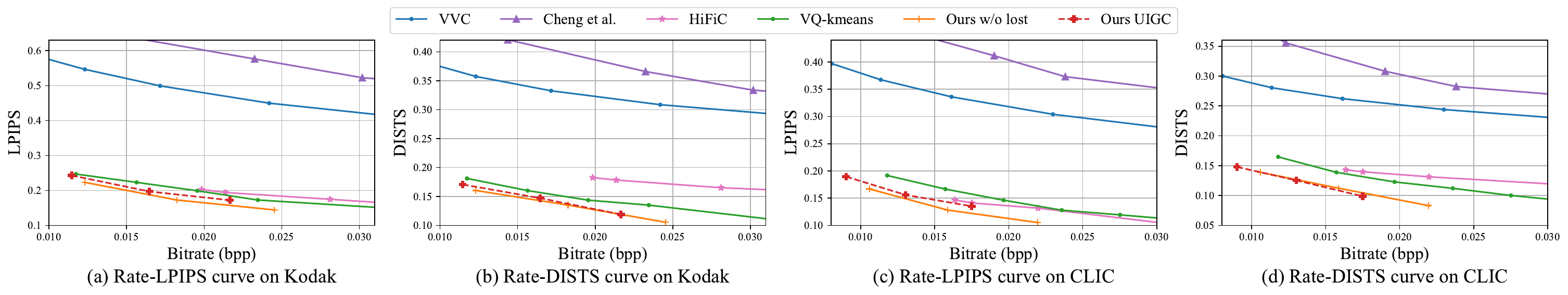}
    \caption{\textbf{R-D curves on the Kodak \cite{kodak} and the CLIC \cite{CLIC2020} datasets.} Ours w/o lost: ours without token lost and generation.}
    \label{fig:rd_curve}
\end{figure*}

\subsection{Multi-Stage Transformer}
\label{sec:mst}
As illustrated in \figref{transformer}(a), the transformer in VQGAN \cite{Esser_2021_CVPR} employs a sliding window for raster order autoregressive encoding to manage memory usage. 
However, this design limits the current encoding position to consider only its \emph{upper-left} context for prediction, potentially compromising the accuracy of prior modeling. 
Recognizing the spatial correlation dependencies in images, we introduce the MST inspired by the multi-stage grouping algorithm in \cite{lu2022highefficiency} to enhance this accuracy by rearranging the autoregressive order.
In particular, the token map is partitioned into four groups, and each group undergoes processing in raster order using a sliding window, as presented in 
\figref{transformer}(a).

Consequently, the MST is structured into four distinct stages.
In Stage $0$, tokens in Group $0$ are sequentially encoded, with each token referencing only its upper-left content within the group.
Following this, Stage $1$ encodes tokens in Group $1$, allowing each token to reference surrounding tokens in Group $0$ and upper-left tokens in Group $1$.
This sequential encoding pattern persists in Stages $2$ and $3$, enabling each token to reference surrounding tokens in preceding groups and upper-left tokens in the current group.
Each group of tokens in the sliding window is flattened in raster order and inputted into each stage of the transformer, as depicted in \figref{transformer}(b).
We utilize a GPT-style transformer for autoregressive encoding, aiming to maximize the log$_2$-likelihood of tokens defined as:
\begin{equation}
    L_\text{transformer} = \mathbb{E}_{I \sim p(I)}[\sum\nolimits_{k=0}^{h\times w} - \log_{2} p({z}_{k}|{z}^{m}_{\le k})].
    \label{eq:cross_entropy}
\end{equation}
During training, a random mask is applied to groups $2$-$3$ to simulate lost tokens, and the transformer estimates the categorical distribution as the prior.

%%%%%%%%%%%%%%%%%%%%%%%%%%%%%%%%%%%%%%%%%%%%%%%%%%%%%%%%%%%%%%%%%%%%%%%%%%
% Mask module
% the idea of mask: utilize generation, remove redundancy
% where to mask? gives example: smooth area, random texture
% describe our design
%%%%%%%%%%%%%%%%%%%%%%%%%%%%%%%%%%%%%%%%%%%%%%%%%%%%%%%%%%%%%%%%%%%%%%%%%%
\subsection{Edge-Preserved Checkerboard Mask}
\label{sec:mask}
Another essential concern is the strategic discarding of redundant tokens to achieve bitrate savings without compromising image quality.
In our proposed MST, tokens in Group $0$ and Group $1$ serve as anchors, providing surrounding references for all tokens in the subsequent groups.
Hence, a \emph{checkerboard pattern} is used as the mask template, retaining all tokens in both groups to ensure accurate prior modeling.
Furthermore, tokens associated with the object structure are preserved, recognizing their crucial role in visual perception.
As such, in this section, we propose an edge-preserving checkerboard mask mechanism as demonstrated in \figref{overview}(b).
First, we extract the object structure using the edge extractor \cite{Xie_ICCV_2015}, regarding it as the preserved region $\boldsymbol{R} \in \{ 0,1 \}^{h \times w}$.
Then, our proposed mask module $\boldsymbol{M}$ generates the final mask $\boldsymbol{m} = \boldsymbol{m_{t}} \lor \boldsymbol{R}$, where the $\boldsymbol{m_{t}} \in \{ 0,1 \}^{h \times w}$ represents the checkerboard template.
The positions of Group $0$ and Group $1$ are set to $1$ in the template $\boldsymbol{m_{t}}$ to retain the corresponding tokens.
Our supplementary material provides further details.

%%%%%%%%%%%%%%%%%%%%%%%%%%%%%%%%%%%%%%%%%%%%%%%%%%%%%%%%%%%%%%%%%%%%%%%%%%
% experiment part
%%%%%%%%%%%%%%%%%%%%%%%%%%%%%%%%%%%%%%%%%%%%%%%%%%%%%%%%%%%%%%%%%%%%%%%%%%
\section{Experiments}

%%%%%%%%%%%%%%%%%%%%%%%%%%%%%%%%%%%%%%%%%%%%%%%%%%%%%%%%%%%%%%%%%%%%%%%%%%
% describe for settings
% Datasets. datasets for training and evaluation
% Implementation details. model architecture, training details
% Evaluation Metrics. PSNR, MS-SSIM, LPIPS, DISTS
%%%%%%%%%%%%%%%%%%%%%%%%%%%%%%%%%%%%%%%%%%%%%%%%%%%%%%%%%%%%%%%%%%%%%%%%%%
\subsection{Implementation Details}
We employ the architecture of encoder and decoder from VQGAN \cite{Esser_2021_CVPR}, and utilize the K-means clustering method detailed in \cite{mao2023extreme} to fine-tune the officially provided pre-trained model with a codebook size of 16384, yielding models with codebook sizes of $\{16,64,256\}$ (denoted as VQ16, VQ64, and VQ256) suitable for ultra-low bitrates. 
During the VQ-codec training, we utilize the default settings and training losses as in \cite{Esser_2021_CVPR}.
For the MST, we set the size of the sliding window at $18\times18$.
We train the proposed model on the ImageNet dataset\cite{deng2009imagenet}.
To evaluate the performance of the proposed model, we use two widely recognized datasets in image compression: the Kodak \cite{kodak} and the CLIC datasets \cite{CLIC2020}.

\begin{table}[!t]
\caption{
\textbf{Average BD-LPIPS↓/DISTS↓ gains on the Kodak \cite{kodak} and the CLIC \cite{CLIC2020} datasets.} Anchor: VVC \cite{vvc}.
}
\small
\vspace{-5mm}
\label{tab:bd-distortion}
\begin{center}
\begin{tabular}{lcccc}
\hline
\multicolumn{1}{c}{\multirow{2}{*}{Method}} & \multicolumn{2}{c}{Kodak}                               & \multicolumn{2}{c}{CLIC}                                \\
\multicolumn{1}{c}{}                        & LPIPS                      & DISTS                      & LPIPS                      & DISTS                      \\ \hline
Cheng \textit{et al.} \cite{Cheng_2020_CVPR}                                   & 0.041                      & 0.044                      & 0.043                      & 0.054                      \\
HiFiC\cite{NEURIPS2020_8a50bae2}                                       & -0.260                     & -0.134                     & -0.176                     & -0.113                     \\
VQ-kmeans \cite{mao2023extreme}                                      & \multicolumn{1}{l}{-0.288} & \multicolumn{1}{l}{-0.182} & \multicolumn{1}{l}{-0.170} & \multicolumn{1}{l}{-0.127} \\
Ours w/o lost                               & \multicolumn{1}{l}{-0.321} & \multicolumn{1}{l}{-0.199} & \multicolumn{1}{l}{-0.207} & \multicolumn{1}{l}{-0.151} \\
Ours UIGC                                   & \multicolumn{1}{l}{-0.310} & \multicolumn{1}{l}{-0.195} & \multicolumn{1}{l}{-0.198} & \multicolumn{1}{l}{-0.150} \\ \hline
\end{tabular}
\end{center}
\vspace{-5mm}
\end{table}

%%%%%%%%%%%%%%%%%%%%%%%%%%%%%%%%%%%%%%%%%%%%%%%%%%%%%%%%%%%%%%%%%%%%%%%%%%
% R-D performance
% compared methods. 
% quantitative evaluation
% qualitative evaluation
%%%%%%%%%%%%%%%%%%%%%%%%%%%%%%%%%%%%%%%%%%%%%%%%%%%%%%%%%%%%%%%%%%%%%%%%%%
\subsection{Compression Performance Evaluation}

\noindent \textbf{Compared Methods.}
To assess the effectiveness of our proposed framework, we conduct a benchmark against both traditional standard and neural-based compression frameworks, including the latest VVC \cite{vvc} codec, the typical end-to-end learning-based approach by Cheng \etal\cite{Cheng_2020_CVPR} (MS-SSIM optimized, for better perceptual quality), and generative image compression codecs, such as HiFiC \cite{NEURIPS2020_8a50bae2}, as well as the VQGAN-based codec VQ-Kmeans \cite{mao2023extreme}.
Furthermore, we develop an additional baseline denoted as ``Ours w/o lost'', which directly applies the MST for entropy estimation without lost tokens and prediction.

\begin{figure}[!t]
    \centering
    \includegraphics[width=1.0\linewidth]{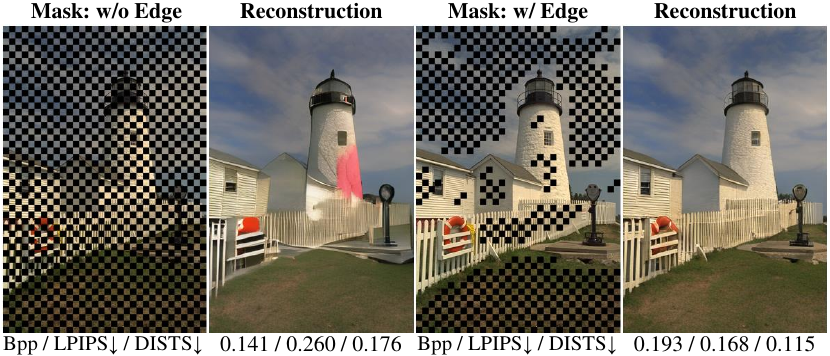}
    \caption{\textbf{Visual comparisons between checkerboard mask with and without edge preservation.}}
    \label{fig:visual_ablation_mask}
\end{figure}

\begin{table}[!t]
\caption{\textbf{R-D performance of RT and MST on the Kodak dataset \cite{kodak}.}  The codec is VQ16 and VQ256.}
%\vspace{-5 mm}
\small
\label{tab:ablation_transformer}
\begin{center}
    \begin{tabular}{lccc}
    \hline
    Method    & Bpp    & LPIPS↓           & DISTS↓           \\ \hline
    VQ16 RT   & 0.0113 & 0.2540          & 0.1712          \\
    VQ16 MST  & 0.0115 & \textbf{0.2429} & \textbf{0.1707} \\
    VQ256 RT  & 0.0216 & 0.1895          & 0.1226          \\
    VQ256 MST & 0.0217 & \textbf{0.1720} & \textbf{0.1189} \\ \hline
    \end{tabular}
\end{center}
\vspace{-5 mm}
\end{table}

\noindent \textbf{Quantitative Evaluation.}
Rather than relying on traditional objective quality assessments like PSNR and SSIM, we incorporate recent perceptual quality-based metrics, such as Learned Perceptual Image Patch Similarity (LPIPS) and Deep Image Structure and Texture Similarity (DISTS), as they offer closer alignment with human perception of images.
Additionally, we employ bits per pixel (bpp) as a metric to evaluate the rate performance. 
We present the R-D performance in \figref{rd_curve}.
We also evaluate the R-D performance improvement using VVC as an anchor with
 Bjontegaard-Delta metric \cite{bjontegaard2001calculation}. %
In particular, we adopt BD-LPIPS and BD-DISTS metrics in \tabref{bd-distortion}, which represent the average perceptual quality improvement under the equivalent bitrate.
It can be clearly observed that our proposed UIGC exhibits superior R-D performance, delivering enhanced visual quality in ultra-low bitrate scenarios (bpp $\leq 0.03$ ).
Note that VQ-kmeans \cite{mao2023extreme} utilizes the same VQ codec as ours. However, its R-D performance is inferior due to the absence of entropy estimation. This finding underscores the effectiveness of the UIGC method, which integrates both entropy estimation and content generation through the use of the prior distribution.
While there is a slight performance drop compared to ``Ours w/o lost" due to the generated content being slightly different from the real image, the UIGC framework further reduces bitrate and achieves comparable perceptual quality as shown in \figref{visual_kodak}.
\noindent \textbf{Qualitative Evaluation.}
\figref{visual_kodak} shows the reconstruction results of various methods on the Kodak dataset \cite{kodak}, with the corresponding bpp, LPIPS, and DISTS. 
At ultra-low bitrates, VVC\cite{vvc} and Cheng \etal\cite{Cheng_2020_CVPR}  exhibit severe blurring. 
Although HiFiC \cite{NEURIPS2020_8a50bae2} and VQ-Kmeans \cite{mao2023extreme} have improved image quality, the detail is not satisfactory:
HiFiC shows the grid artifact on the sea and the abnormal reconstruction of the parrot's eye; VQ-Kmeans produces distorted structure on the boat.
In contrast, our methods exhibit more natural sea surface and object structures (\eg the boat, parrots, trees).
Moreover, the UIGC further saves bitrate while maintaining almost the same visual quality as ``Ours w/o lost", with only a negligible loss in water texture details.

\begin{figure} [!t]
    \centering
    \includegraphics[width=1.0\linewidth]{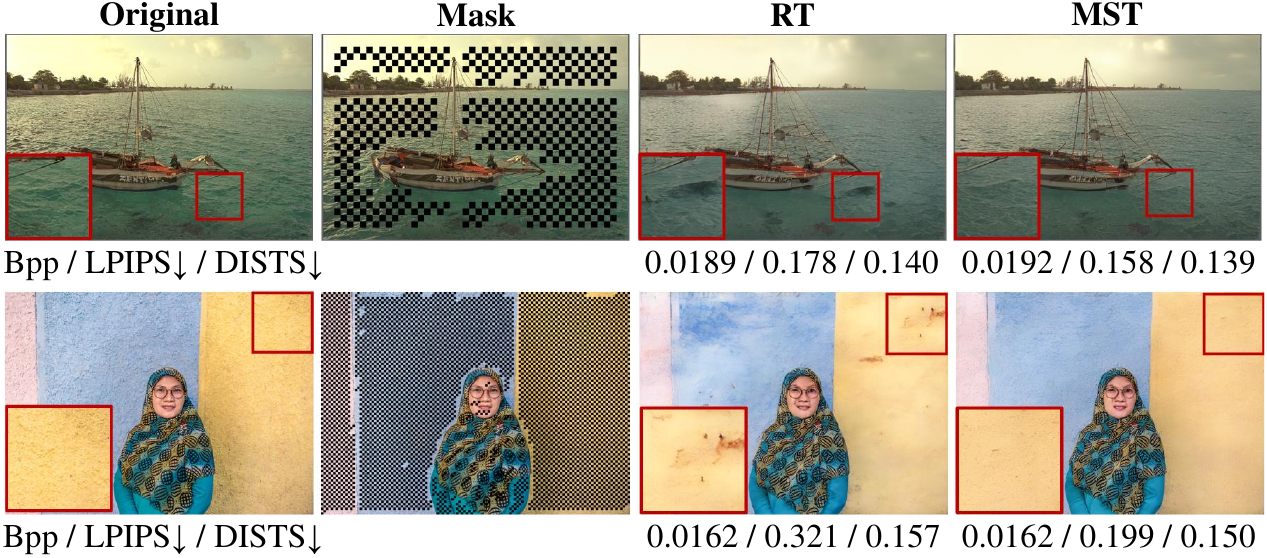}
    \caption{\textbf{Visual Comparisons between MST and RT.}}
    \label{fig:visual_ablation}
    %\vspace{-3 mm}
\end{figure}

\begin{figure} [!t]
    \centering
    \includegraphics[width=1.0\linewidth]{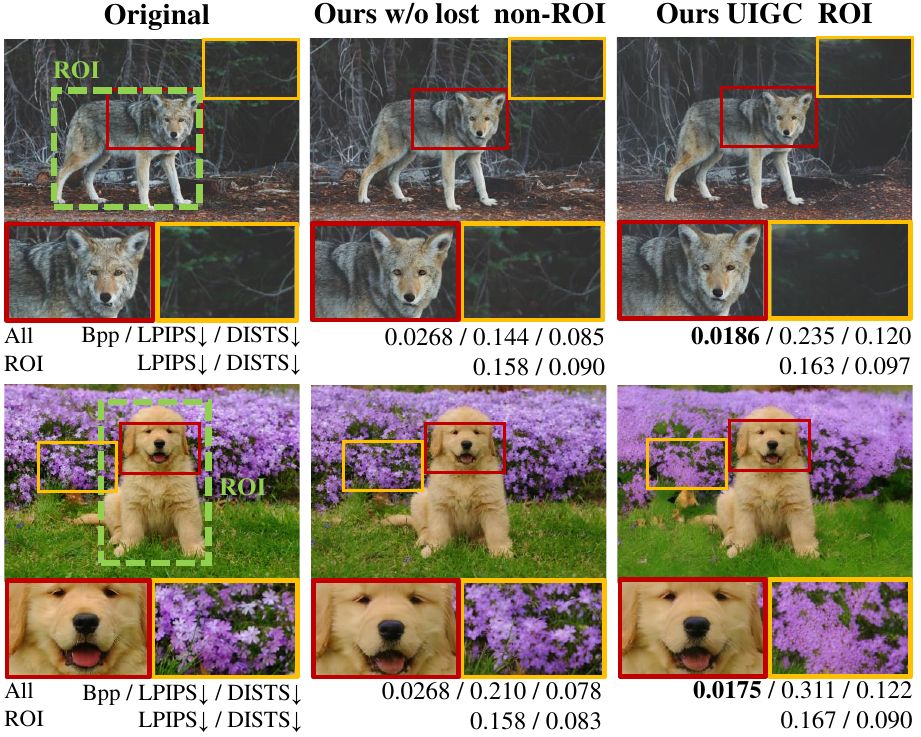}
    \caption{\textbf{Examples of ROI coding.} ``Ours UIGC ROI" ensures ROI quality while significantly lowering bitrate.}
    \label{fig:visual_ROI}
\end{figure}

\subsection{Ablation Studies and Discussion}
\noindent \textbf{Efficiency of the MST.}
We conduct experiments to ascertain the advancements of the proposed MST over the Raster Transformer (RT) detailed in \cite{Esser_2021_CVPR}.
\tabref{ablation_transformer} indicates that while RT and MST exhibit similar levels of entropy coding efficiency, the MST surpasses RT in the perceptual quality of the generated images.
\figref{visual_ablation} presents examples of images generated by both MST and RT. The RT tends to create unnatural textures in areas like the sea and walls due to its predominant reliance on upper-left positional references, resulting in a lower visual quality compared to MST.

\noindent \textbf{Mask Pattern.}
To evaluate the importance of tokens associated with the object structure, we test two mask patterns: checkerboard with and without edge preservation.
\figref{visual_ablation_mask} demonstrates that although excluding edge tokens contributes to further bitrate reduction, it simultaneously causes issues like distorted edges and abnormal content (for instance, the red region on the tower). This observation effectively confirms the essentiality of implementing an edge preservation mechanism in our approach.

\noindent \textbf{Region of Interest Compression.}
Region of interest (ROI) coding, essential in multimedia applications, requires high-quality compression of selected regions while allowing for more aggressive compression in non-essential areas to reduce bitrate. 
Our proposed UIGC framework is adept at accommodating this need by selectively preserving tokens in the ROI.
As shown in \figref{visual_ROI}, UIGC's approach to ROI coding not only significantly reduces the bitrate but also maintains an aesthetically pleasing visual quality in the regions of interest.

\section{conclusions}
In this work, we propose a novel UIGC paradigm, specifically tailored for ultra-low bitrate image compression. This versatile framework adeptly utilizes the learned prior distribution for both entropy estimation and the regeneration of lost tokens.
We further design the MST to boost prior modeling accuracy, and introduce an edge-preserving checkerboard mask pattern to discard unnecessary tokens for bitrate saving.
Our experimental results validate the UIGC's superiority over existing codecs in visual quality, particularly in ultra-low bitrate ($\leq$ 0.03 bpp) scenarios.
We believe that the UIGC scheme represents a significant advancement in generative compression, charting a new course for future developments.
\small
\bibliographystyle{IEEEbib}
\bibliography{MaskVQGAN}

\begin{thebibliography}{10}

\bibitem{vvc}
Benjamin Bross, Ye-Kui Wang, Yan Ye, Shan Liu, Jianle Chen, Gary~J. Sullivan, and Jens-Rainer Ohm,
\newblock ``Overview of the versatile video coding (vvc) standard and its applications,''
\newblock {\em TCSVT}, 2021.

\bibitem{NEURIPS2018_53edebc5}
David Minnen, Johannes Ball\'{e}, and George~D Toderici,
\newblock ``Joint autoregressive and hierarchical priors for learned image compression,''
\newblock in {\em NIPS}, 2018.

\bibitem{Cheng_2020_CVPR}
Zhengxue Cheng, Heming Sun, Masaru Takeuchi, and Jiro Katto,
\newblock ``Learned image compression with discretized gaussian mixture likelihoods and attention modules,''
\newblock in {\em CVPR}, June 2020.

\bibitem{He_2021_CVPR_checkerboard}
Dailan He, Yaoyan Zheng, Baocheng Sun, Yan Wang, and Hongwei Qin,
\newblock ``Checkerboard context model for efficient learned image compression,''
\newblock in {\em CVPR}, 2021.

\bibitem{lu2022highefficiency}
Ming Lu, Fangdong Chen, Shiliang Pu, and Zhan Ma,
\newblock ``High-efficiency lossy image coding through adaptive neighborhood information aggregation,''
\newblock {\em arXiv preprint}, 2022.

\bibitem{Agustsson_2019_ICCV}
Eirikur Agustsson, Michael Tschannen, Fabian Mentzer, Radu Timofte, and Luc~Van Gool,
\newblock ``Generative adversarial networks for extreme learned image compression,''
\newblock in {\em ICCV}, 2019.

\bibitem{NEURIPS2020_8a50bae2}
Fabian Mentzer, George~D Toderici, Michael Tschannen, and Eirikur Agustsson,
\newblock ``High-fidelity generative image compression,''
\newblock in {\em NIPS}, 2020.

\bibitem{fcc}
Shoma Iwai, Tomo Miyazaki, Yoshihiro Sugaya, and Shinichiro Omachi,
\newblock ``Fidelity-controllable extreme image compression with generative adversarial networks,''
\newblock in {\em ICPR}, 2021.

\bibitem{Agustsson_2023_CVPR}
Eirikur Agustsson, David Minnen, George Toderici, and Fabian Mentzer,
\newblock ``Multi-realism image compression with a conditional generator,''
\newblock in {\em CVPR}, 2023.

\bibitem{mao_layerd}
Jianhui Chang, Qi~Mao, Zhenghui Zhao, Shanshe Wang, Shiqi Wang, Hong Zhu, and Siwei Ma,
\newblock ``Layered conceptual image compression via deep semantic synthesis,''
\newblock in {\em ICIP}, 2019.

\bibitem{mao_conceptual}
Jianhui Chang, Zhenghui Zhao, Chuanmin Jia, Shiqi Wang, Lingbo Yang, Qi~Mao, Jian Zhang, and Siwei Ma,
\newblock ``Conceptual compression via deep structure and texture synthesis,''
\newblock {\em TIP}, 2022.

\bibitem{2021thousand_to_one}
Jianhui Chang, Zhenghui Zhao, Lingbo Yang, Chuanmin Jia, Jian Zhang, and Siwei Ma,
\newblock ``Thousand to one: Semantic prior modeling for conceptual coding,''
\newblock in {\em ICME}, 2021.

\bibitem{chang2023semantic}
Jianhui Chang, Jian Zhang, Jiguo Li, Shiqi Wang, Qi~Mao, Chuanmin Jia, Siwei Ma, and Wen Gao,
\newblock ``Semantic-aware visual decomposition for image coding,''
\newblock {\em IJCV}, 2023.

\bibitem{mao2023extreme}
Qi~Mao, Tinghan Yang, Yinuo Zhang, Zijian Wang, Meng Wang, Shiqi Wang, and Siwei Ma,
\newblock ``Extreme image compression using fine-tuned vqgans,''
\newblock {\em arXiv preprint}, 2023.

\bibitem{kingma2022autoencoding}
Diederik~P Kingma and Max Welling,
\newblock ``Auto-encoding variational bayes,''
\newblock {\em arXiv preprint}, 2013.

\bibitem{goodfellow2014generative}
Ian Goodfellow, Jean Pouget-Abadie, Mehdi Mirza, Bing Xu, David Warde-Farley, Sherjil Ozair, Aaron Courville, and Yoshua Bengio,
\newblock ``Generative adversarial nets,''
\newblock {\em NIPS}, vol. 27, 2014.

\bibitem{ho2020denoising}
Jonathan Ho, Ajay Jain, and Pieter Abbeel,
\newblock ``Denoising diffusion probabilistic models,''
\newblock {\em NIPS}, 2020.

\bibitem{deletang2023language}
Gr{\'e}goire Del{\'e}tang, Anian Ruoss, Paul-Ambroise Duquenne, Elliot Catt, Tim Genewein, Christopher Mattern, Jordi Grau-Moya, Li~Kevin Wenliang, Matthew Aitchison, Laurent Orseau, et~al.,
\newblock ``Language modeling is compression,''
\newblock {\em arXiv preprint}, 2023.

\bibitem{Esser_2021_CVPR}
Patrick Esser, Robin Rombach, and Bjorn Ommer,
\newblock ``Taming transformers for high-resolution image synthesis,''
\newblock in {\em CVPR}, 2021.

\bibitem{Chang_2022_CVPR}
Huiwen Chang, Han Zhang, Lu~Jiang, Ce~Liu, and William~T. Freeman,
\newblock ``Maskgit: Masked generative image transformer,''
\newblock in {\em CVPR}, 2022.

\bibitem{kodak}
Eastman Kodak,
\newblock ``Kodak photocd dataset,'' 2013.

\bibitem{CLIC2020}
Toderici George, Shi Wenzhe, Timofte Radu, Theis Lucas, Balle Johannes, Agustsson Eirikur, Johnston Nick, and Mentzer Fabian,
\newblock ``Workshop and challenge on learned image compression (clic2020),'' 2020.

\bibitem{Xie_ICCV_2015}
Saining Xie and Zhuowen Tu,
\newblock ``Holistically-nested edge detection,''
\newblock in {\em ICCV}, 2015.

\bibitem{deng2009imagenet}
Jia Deng, Wei Dong, Richard Socher, Li-Jia Li, Kai Li, and Li~Fei-Fei,
\newblock ``Imagenet: A large-scale hierarchical image database,''
\newblock in {\em CVPR}, 2009.

\bibitem{bjontegaard2001calculation}
Gisle Bjontegaard,
\newblock ``Calculation of average psnr differences between rd-curves,''
\newblock {\em ITU SG16 Doc. VCEG-M33}, 2001.

\bibitem{gailly2004zlib}
Jean-loup Gailly and Mark Adler,
\newblock ``Zlib compression library,''
\newblock 2004.

\bibitem{begaint2020compressai}
Jean B{\'e}gaint, Fabien Racap{\'e}, Simon Feltman, and Akshay Pushparaja,
\newblock ``Compressai: a pytorch library and evaluation platform for end-to-end compression research,''
\newblock {\em arXiv preprint arXiv:2011.03029}, 2020.

\end{thebibliography}

\clearpage

\section*{A Implementation Details}
\label{sec:details}

\begin{figure}[t]
    \centering
    \includegraphics[width=1.0\linewidth]{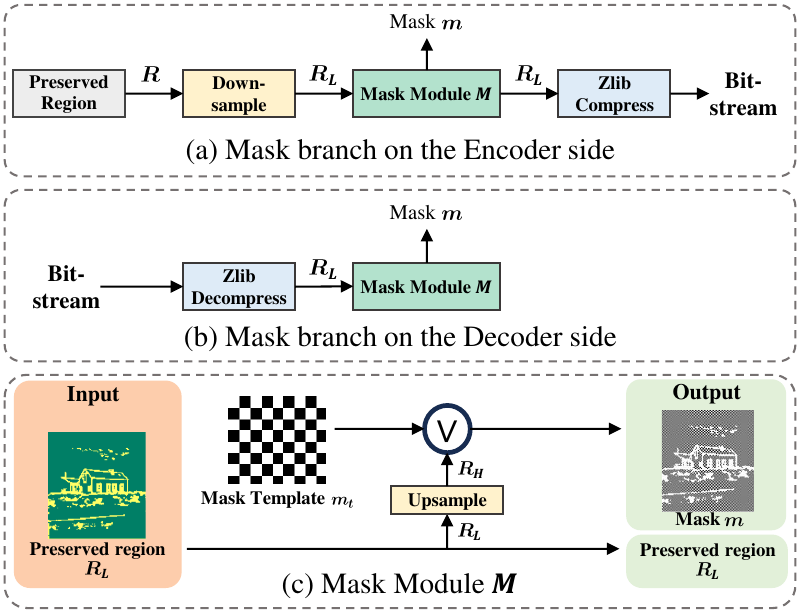}
    \caption{\textbf{Details of the downsampling and upsampling design in our mask mechanism}. (a) the downsampling and compression of the preserved region $\boldsymbol{R}$ in the encoder side, as well as the mask generation; (b) the decompression of the $\boldsymbol{R_L}$ and mask generation in the decoder side; (c) upsampling design inside the Mask Module $\boldsymbol{M}$.}
    \label{fig:mask}
\end{figure}

\subsection*{A.1 Multi-Stage Transformer}
The GPT model used in our Multi-Stage Transformer (MST) consists of 32 layers of transformers. 
Each layer of transformer has 16 heads of self-attention and the embedding dimension is 1152.
During training, image patches of size 288 are randomly cropped. 
These patches are encoded into an 18x18 token map by VQGAN’s Encoder \cite{Esser_2021_CVPR}, matching MST’s sliding window size.
Additionally, masks are randomly applied to the tokens of Group 2 and Group 3 with a variable probability ranging from 0\% to 100\%.

\subsection*{A.2 Edge-Preserved Checkerboard Mask}
In our design, we downsample the preserved region $\boldsymbol{R} \in\{ 0,1 \} ^{h \times w}$ into $\boldsymbol{R_L} \in\{ 0,1 \} ^{0.5h \times 0.5w}$ to reduce bit cost. 
The encoder-side downsampling and compression of $\boldsymbol{R}$ is depicted in \figref{mask}(a), while \figref{mask}(b) demonstrates the decoder-side decompression of $\boldsymbol{R_L}$.
Given that $\boldsymbol{R_L}$ only contains $0$ and $1$, we consider it as a bitstream and apply lossless compression using the Zlib codec \cite{gailly2004zlib}.
Our Mask Module $\boldsymbol{M}$ then upsamples the $\boldsymbol{R_L}$ into $\boldsymbol{R_H} \in\{ 0,1 \} ^{h \times w}$ and generates the final mask $\boldsymbol{m}$, as illustrated in \figref{mask} (c).

\begin{figure*}[t]
    \centering
    \includegraphics[width=1.0\linewidth]{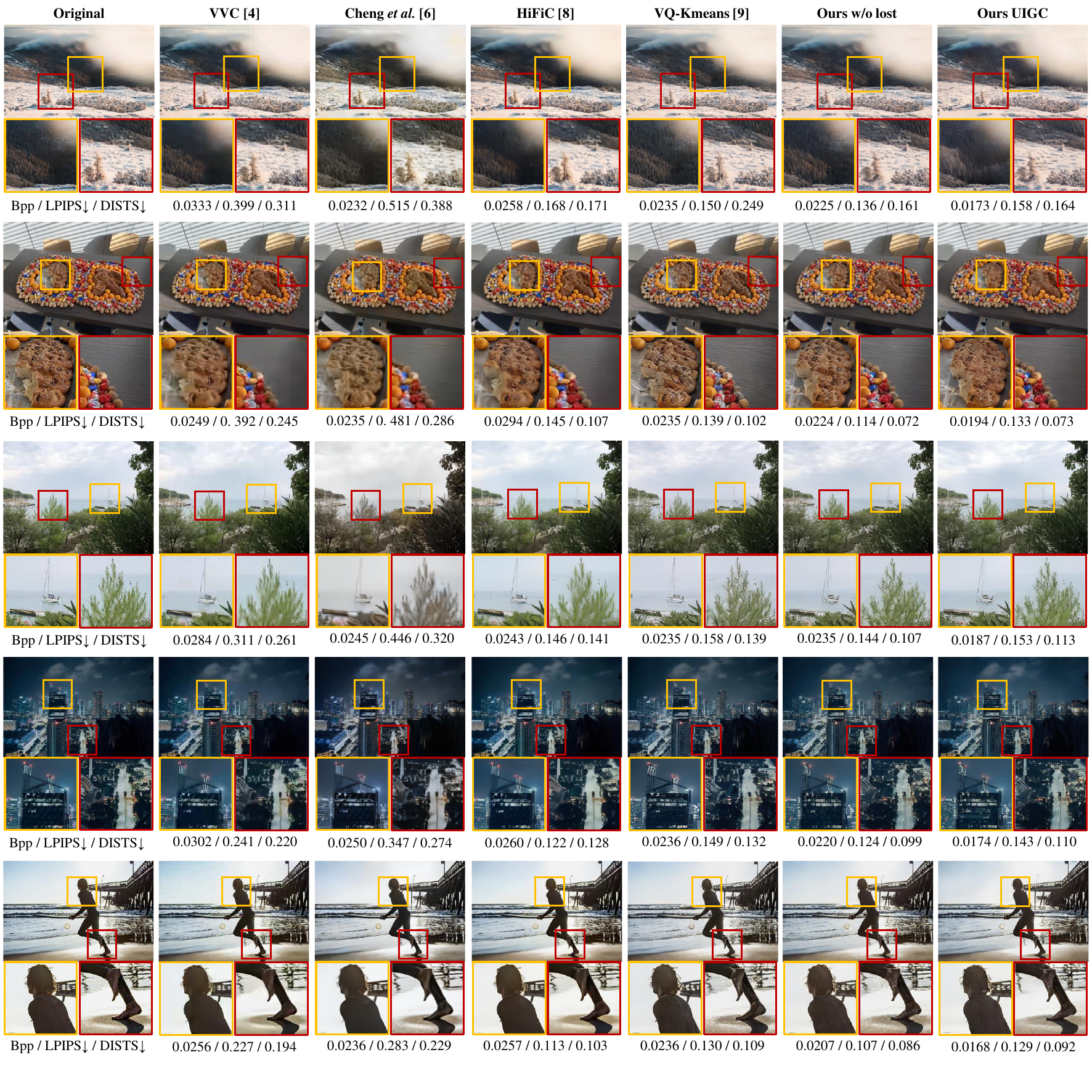}
    \caption{\textbf{Qualitative comparisons on the CLIC dataset \cite{CLIC2020}.} In particular, ↓ indicates that lower is better.}
    \label{fig:visual_CLIC}
\end{figure*}

\section*{B Experimental Details}

\subsection*{B.1 Traditional Codec}
We use the VTM version 22.2 as the implementation of the latest traditional compression standard VVC \cite{vvc}.
The following commands are used for the VVC codec:
\begin{verbatim}
    EncoderApp
        -i {input file name}
        -c encoder_intra_vtm.cfg
        -q {QP}
        -o /dev/null
        -b {bitstream file name}
        -wdt {image width}
        -hgt {image height}
        -fr 1
        -f 1
        --InputChromaFormat=444
        --InputBitDepth=8
        --ConformanceWindowMode=1
    DecoderApp
        -b {bitstream file name}
        -o {reconstruction file name}
        -d 8
\end{verbatim}

\subsection*{B.2 Neural-based Codec}
All neural-based codecs are trained on the ImageNet dataset \cite{deng2009imagenet}.
The implementation details are as follows:
\begin{itemize}
    \item \textbf{Cheng \textit{et al.} \cite{Cheng_2020_CVPR}}: 
    We employ the CompressAI library \cite{begaint2020compressai} implementation, adhering to its default losses and training strategies. 
    The hyperparameter $\lambda$ is adjusted for ultra-low bitrate compression training.

    \item \textbf{HiFiC \cite{NEURIPS2020_8a50bae2}}: 
    We utilize the pytorch implementation of the HiFiC \cite{NEURIPS2020_8a50bae2}, available at \href{https://github.com/Justin-Tan/high-fidelity-generative-compression}{Justin-Tan/high-fidelity-generative-compression}.
    The default losses and training strategies are maintained for model training, with the hyperparameter $\lambda$ adjusted for ultra-low bitrate compression.

    \item \textbf{VQ-kmeans \cite{mao2023extreme}}: 
    This method is implemented based on the pre-trained VQGAN \cite{Esser_2021_CVPR} model with a codebook size of 16384, generating models with codebook sizes of \{8, 16, 32, 64, 128, 256\}. 
    The training settings and losses follow the method described in \cite{mao2023extreme}.
    
\end{itemize}

\section*{C. Visual Results}
We provide CLIC reconstructions for qualitative comparison, as \figref{visual_CLIC} shows.

\end{document}